\documentclass{article}
\usepackage[nonatbib]{neurips_2024}

\usepackage[utf8]{inputenc} % allow utf-8 input
\usepackage[T1]{fontenc}    % use 8-bit T1 fonts
\usepackage{hyperref}       % hyperlinks
\usepackage{url}            % simple URL typesetting
\usepackage{booktabs}       % professional-quality tables
\usepackage{amsfonts}       % blackboard math symbols
\usepackage{nicefrac}       % compact symbols for 1/2, etc.
\usepackage{microtype}      % microtypography
\usepackage{xcolor}         % colors
\usepackage{amsmath}
\usepackage{todonotes}
 
\usepackage{graphicx}
\usepackage{algorithm}
\usepackage{algorithmic}
\usepackage{subcaption} 
\usepackage{soul}
\usepackage{pifont}
\usepackage{wrapfig}
\title{Adapting Segment Anything Model (SAM) to Experimental Datasets via Fine-Tuning on GAN-based Simulation: A Case Study in Additive Manufacturing}
\author{
    Anika Tabassum\textsuperscript{\rm 1}, %\thanks{With help from the AAAI Publications Committee.}\\
    Amirkoushyar Ziabari\textsuperscript{\rm 1}\\
    \textsuperscript{\rm 1}Oak Ridge National Laboratory\\
    \texttt{\{tabassuma,ziabariak\}@ornl.gov}
}

\begin{document}

\maketitle

\begin{abstract}
Industrial X-ray computed tomography (XCT) is a powerful tool for non-destructive characterization of materials and manufactured components. 
XCT commonly accompanied by advanced image analysis and computer vision algorithms to extract relevant information from the images. 
Traditional computer vision models often struggle due to noise, resolution variability, and complex internal structures, particularly in scientific imaging applications. 
State-of-the-art foundational models, like the \emph{Segment Anything Model (SAM)}—designed for general-purpose image segmentation—have revolutionized image segmentation across various domains, yet their application in specialized fields like materials science remains under-explored. 
In this work, we explore the application and limitations of SAM for industrial X-ray CT inspection of additive manufacturing components. 
We demonstrate that while \emph{SAM}  shows promise, it struggles with out-of-distribution data, multiclass segmentation, and computational efficiency during fine-tuning. 
To address these issues, we propose a fine-tuning strategy utilizing parameter-efficient techniques, specifically \emph{Conv-LoRa} , to adapt \emph{SAM}  for material-specific datasets. 
Additionally, we leverage generative adversarial network (GAN)-generated data to enhance the training process and improve the model's segmentation performance on complex X-ray CT data.
Our experimental results highlight the importance of tailored segmentation models for accurate inspection, showing that fine-tuning \emph{SAM}  on domain-specific scientific imaging data significantly improves performance. 
However, despite improvements, the model's ability to generalize across diverse datasets remains limited, highlighting the need for further research into robust, scalable solutions for domain-specific segmentation tasks. 
Code and data are available for research purposes~\footnote{https://github.com/anikat1/SAM-Material-GAN}.

\end{abstract}

\section{Introduction}\label{sec:intro}%\vspace{-0.35cm}
\begin{figure*}[!htbp]
\centering %, bb=0 0 900 1000
\includegraphics[width=0.98\textwidth]{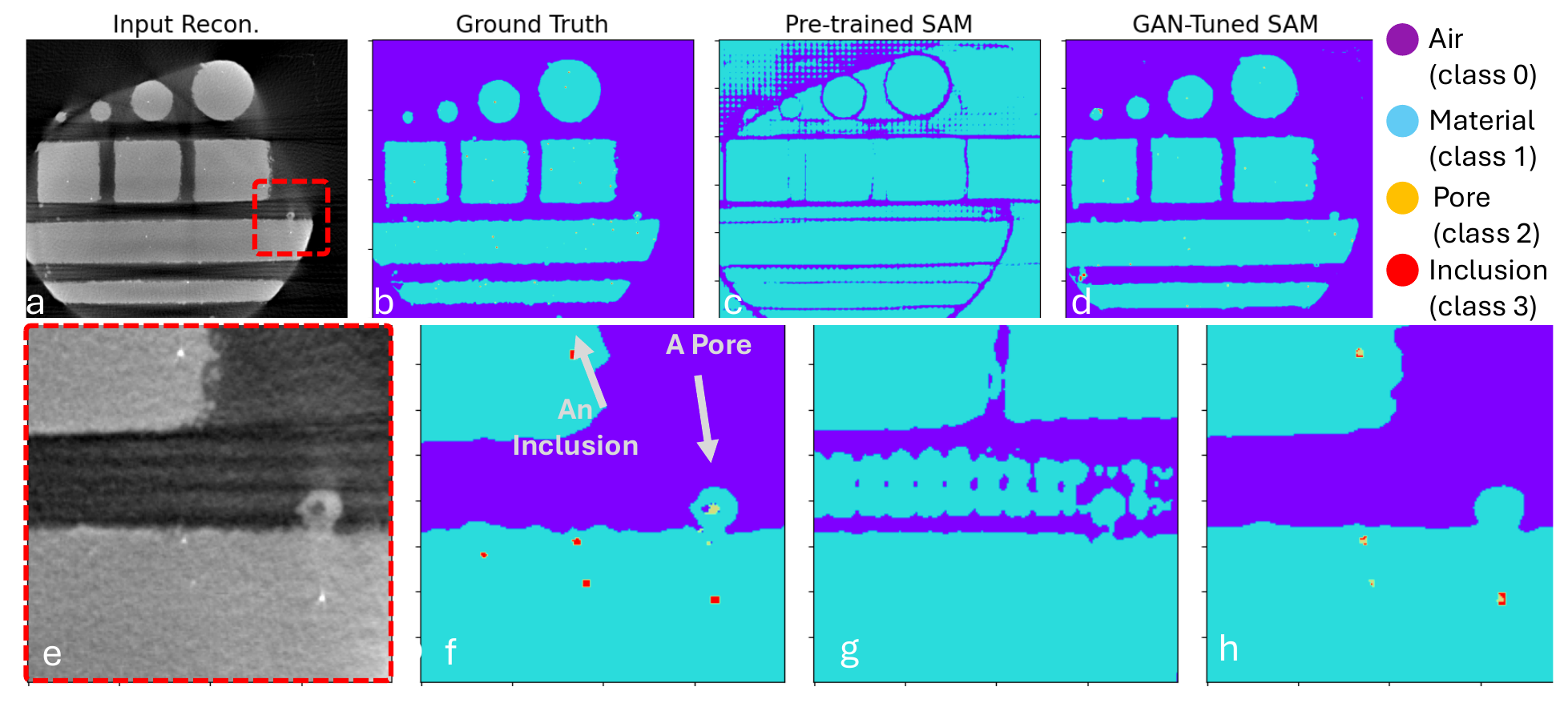}
\caption{An example cross section (slice) comparison from the 3D volume of an additive manufacturing (\emph{AM} ) component. 
a) Experimental XCT reconstruction; 
b) Ground Truth Segmentation (obtained through a higher resolution, better quality scan of the same object that are quire expensive (time and cost) to acquire every time). Four classes (air, material, pore, and inclusion) are identified in the ground truth image. 
Predictions obtained by c) pretrained, and d) finetuned \emph{SAM} . 
An expanded view of the ROI (red dashed box in panel a), for each of the panels a-d are shown in e-h, respectively. }
% (c) Zero-shot Segmentation obtained from \emph{SAM}  (d) Segmentation obtained after fine-tuning \emph{SAM}  on GAN-based \emph{AM}  data. Fig.(e)-(f) shows the similar output on a snippet of this sample image data. }
\label{fig:intro}
%\vspace{-.5cm}
\end{figure*}
X-ray computed tomography (XCT) has become an indispensable tool for three-dimensional (3D) non-destructive evaluation and characterization (NDE/NDC) in scientific imaging applications. 
XCT enables detailed 3D visualization and analysis of the internal structure of materials and components, providing valuable insights into microstructures, defect detection, and material behavior over time. 
Industries such as aerospace, automotive, and energy rely heavily on XCT for quality control, process optimization, and research purposes, as it allows for precise inspection of complex components without physical alteration. 
In the field of additive manufacturing (\emph{AM} ), XCT is widely used for NDE to detect flaws that may form during the build process. 
Additive manufacturing (also known as 3D printing) involves building parts layer by layer, enabling the creation of complex and customized structures. 
However, this process can introduce defects, such as voids or cracks, which affect the final part's strength and reliability. 
XCT allows for the non-destructive identification and detection of these flaws, enabling qualification of parts for use and optimizing manufacturing processes by adjusting parameters such as laser power or speed to minimize defects~\cite{kim2019influence,sundar2021flaw,ziabari2023enabling,kozjek2023iterative}.

Traditional segmentation algorithms developed for defect detection in XCT images often require extensive manual work and rely on specific parameters that are highly dependent on data quality~\cite{du2019laboratory,villarraga2024non}. 
These methods are often inconsistent across different materials and defect types, limiting their scalability for broader applications. 
Additionally, their lack of adaptability reduces their effectiveness in diverse manufacturing scenarios~\cite{RAHMAN2024114317}.

Supervised deep learning (DL) methods have been developed to improve characterization and segmentation quality~\cite{deshpande2024deep,wong2020automatic,gobert2020porosity}. 
However, these methods require significant amounts of labeled data and often lack generalization capability across datasets. 
To enhance generalizability and reduce reliance on labeled data, unsupervised and weakly-supervised DL-based algorithms have been developed~\cite{fuchs2019self,nemati2024self,zang2021intratomo,wang2023contrastive}. 
These techniques, typically trained on large, diverse image sets, can be fine-tuned to segment new images not present in the training data, with few/zero-shot learning.

Recently, foundational segmentation models have emerged, offering greater flexibility and performance. 
These models are designed to automatically and robustly delineate different material phases within microscopic images, regardless of material type, image resolution, or imaging source. 
Leveraging such advanced models significantly enhances the precision and efficiency of analysis, enabling advancements in materials design and quality control. 

One such foundational model is the Segment Anything Model (\emph{SAM}), pre-trained on multiple publicly available ImageNet datasets~\cite{kirillov2023segment}. 
Fine-tuning \emph{SAM}  has shown promising results on medical images~\cite{mazurowski2023segment} and industrial CT images~\cite{weinberger2024unsupervised}. 
However, these datasets typically involve single-class semantic segmentation, or \emph{SAM}  was fine-tuned with different prompts that augment the training dataset. 
Additionally, early experiments with \emph{SAM}  on segmenting small flaws and anomalies in XCT data from materials have shown limitations. 
As shown in Fig.~\ref{fig:intro}(c), the pre-trained \emph{SAM}  model struggles to differentiate between pore and inclusion structures (dark and bright spot anomalies) in materials XCT images.

In this work, we aim to address these limitations by proposing a fine-tuning strategy utilizing parameter-efficient techniques, such as \emph{Conv-LoRa} ~\cite{zhong2024convolution}, to adapt \emph{SAM}  for material-specific datasets. 
Additionally, we leverage GAN-generated data to enhance the training process and improve the model's segmentation performance on challenging XCT data used for inspecting additive manufacturing parts. 
Our approach demonstrates significant improvements in the segmentation of X-ray computed tomography (XCT) images, particularly in handling out-of-distribution data and fine-tuning for specialized applications.
In addition, we address some of the challenges encountered during fine-tuning and inference of \emph{SAM}  on this data. 
It is worth noting that we selected a case study in additive manufacturing, as the anomalies and flaws in \emph{AM}  parts' XCT images result in an extremely unbalanced segmentation problem. 
Additionally, automated high-throughput inspection of \emph{AM}  parts has significant implications for the next generation of manufacturing processes, addressing supply chain shortages and enabling Industry 4.0.

\vspace{-0.25cm}

\section{Data Generation}\label{sec:data}%\vspace{-0.25cm}
\subsection{CycleGAN for Unpaired Data Generation and Domain Adaptation} %\vspace{-0.2cm}
CycleGAN is a well-established method for unpaired image-to-image translation and is widely used for domain adaptation tasks~\cite{zhu2017unpaired}. 
CycleGAN learns two mappings: one from domain A to domain B, and a reverse mapping from B to A. Its cycle consistency loss ensures that, when an image is mapped from one domain to another and then back, it closely resembles the original image:

\begin{equation}
    L_{\text{cyc}}(G, F) = \mathbb{E}_{x \sim p_{\text{data}}(x)}[\| F(G(x)) - x \|_1] + \mathbb{E}_{y \sim p_{\text{data}}(y)}[\| G(F(y)) - y \|_1]
\end{equation}

where \( G \) and \( F \) are the generators for mapping between the two domains. This cycle consistency property allows the model to effectively learn mappings between unpaired datasets without requiring one-to-one correspondence.
In \cite{ziabari2022simurgh}, authors utilized computer-aided design (CAD) models of metal additive manufacturing (AM) parts to simulate realistic X-ray CT data. A library of flaws—typically appearing as dark spots within the material regions—was embedded into the CAD models to generate datasets incorporating realistic defect distributions. These flaws were simulated using a physics-based X-ray CT simulator, modeling beam hardening and noise characteristics inherent to the XCT process. Due to significant differences between simulated and real XCT data, including noise and artifacts, the models faced challenges in accurately mimicking real-world data conditions. To address this domain shift, a CycleGAN-based domain adaptation approach was employed to generate realistic datasets. These enhanced datasets were then used to train deep learning models for improved XCT reconstruction by suppressing noise and artifacts.

In comparison to \cite{ziabari2022simurgh}, our work expands the data generation process to include not only material flaws but also additional anomalies known as \textit{inclusions}, which typically consist of denser materials and appear as bright spots in industrial XCT images. 
A known challenge with CycleGANs is the difficulty in preserving small features such as flaws. 
In our experiments, pores generated by CycleGAN tended to have higher contrast and less blurriness compared to real data, affecting their realism. To address this, we ensured that the attenuation coefficient distributions for flaws in our simulations closely matched those obtained from real data, while other attenuation coefficients were adjusted based on the density of the materials.

Another important factor addressed in our methodology was the alignment of simulated and real data. Given that unpaired datasets do not guarantee perfect registration, we ensured that the orientations of the simulated data roughly matched those of the real samples on the XCT stage. This alignment helped mitigate some limitations of CycleGAN in maintaining the global structure of objects during domain adaptation, improving the overall quality of the generated data without needing perfect registration.
It is worth noting that we also experimented with constraining the loss function of CycleGAN (similar to SP-CycleGAN~\cite{joon2017nuclei}), as well as Constrained Unpaired Translation GAN (CUT-GAN)~\cite{park2020contrastive} to further improve feature preservation, but the improvement was marginal, if any. While the core CycleGAN architecture was sufficient for much of the domain adaptation task, further fine-tuning may be needed for handling the nuances of complex material flaws.

For training, we used 1746 images with a resolution of 1200x1200 pixels, with a pixel size of $17.3\times17.3 \mu m^2$. We utilized full-resolution image patches of 900x900 to maintain feature fidelity during training (the largest crop size we could keep with GPU constraints that was divisible by 4). The model was initialized with a learning rate of 0.0002, which was reduced after 100 epochs. Training was conducted for 200 epochs in total, with the first 100 epochs using the fixed learning rate and the remaining 100 epochs applying a linear decay. The batch size was set to 4, and a pool size of 200 was used to store generated images for reuse in the discriminator to prevent overfitting. The pool size, set empirically, helps generalize by exposing the discriminator to a broader range of images, thus stabilizing training.

Training was performed on four A100 Nvidia GPUs, each with 40GB of memory, and model weights were saved every two epochs. We evaluated the performance of the models using the Fréchet Inception Distance (FID) and Structural Similarity Index Measure (SSIM) between the real XCT data and the CycleGAN-generated volumes. These metrics allowed us to select the best-performing model for further analysis.
We trained $3$ separate models to capture a range of material densities and measurement settings, ensuring that our models generalized across different materials. Once training was completed, we generated 16 volumetric datasets: $8$ for Material $1$, $4$ for Material $2$, and $4$ for Material $3$. From these, one volume from Material $1$ was used to train a U-Net segmentation model, while one volume each from Material $1$ and Material $2$ was used to train a Segment Anything Model (SAM). One volume from Material $3$ was set aside as OoD synthetic data for testing.
\vspace{-0.4cm}
\subsection{Real Data Preparation}%\vspace{-0.2cm}

X-ray Computed Tomography (XCT) is a technique where an object is scanned from multiple angles, with each angle producing a projection recorded on a detector. These projections are then processed using a reconstruction algorithm to generate a 3D representation of the object. In industrial applications, especially when dealing with large detectors and the need to capture fine features, thousands of views are often required to achieve high-quality reconstructions. However, this process can be time-consuming, labor-intensive, and costly \cite{VASARHELYI2020100084}. The material's complexity also plays a significant role in the quality of the reconstruction. For instance, dense metallic parts, such as those produced in additive manufacturing, can significantly degrade the quality of the reconstruction as material density increases or the geometry becomes more complex \cite{carmignato2018industrial}.

Additionally, fabrication and manufacturing processes, combined with material properties, may introduce anomalies and artifacts in the XCT data, such as pores, inclusions, and noise. To create both in-distribution (InD) and out-of-distribution (OoD) datasets for testing deep learning and SAM algorithms, we scanned five separate parts from four distinct materials. We selected a few cases that are diverse enough through difference in noise texture (due to materials and scan setting variation), and type of flaws and anomalies present in the data, so that we can evaluate the performance of the algorithm.
Hence, the resulting reconstructions were intentionally varied, producing both noisy and clean datasets, with different types of anomalies and varying numbers of pores. 
For each part, we also conducted a high-resolution reference scan, using an increased number of views and a deep learning-based model-based iterative reconstruction algorithm \cite{rahman2023neural,pyMBIR}. 
This yielded high-quality reconstructions, allowing us to segment pores and inclusions at a higher resolution. These high-quality segmentation were then used as ground truth for validating the lower-quality scans employed for testing our models.

In practical industrial settings, performing these high-resolution reference scans is often impractical or cost-prohibitive, particularly in environments like manufacturing facilities where thousands of samples may need to be scanned for process optimization or part qualification. The high cost of detailed scans results in a trade-off between cost and quality. Consequently, many scans are conducted using sub-optimal, faster, and sparser acquisition settings, which leads to lower-quality reconstructions. This reduction in quality complicates the inspection process, especially in industrial applications. Artifacts arising from sparse data, complex geometries, and high-density materials can further degrade even high-quality scans. Therefore, automated deep learning techniques, particularly those that adapt to parameter variations, become essential for improving the inspection process. By generating both high-quality and low-quality scans, these datasets help develop more robust models capable of handling the variability and complexities inherent in industrial XCT data.
In the following, we describe these real test dataset.

\vspace{-0.25cm}
\subsection{Training and Test Dataset}%\vspace{-0.25cm}

\begin{figure}
    \centering
    \includegraphics[width=1\linewidth]{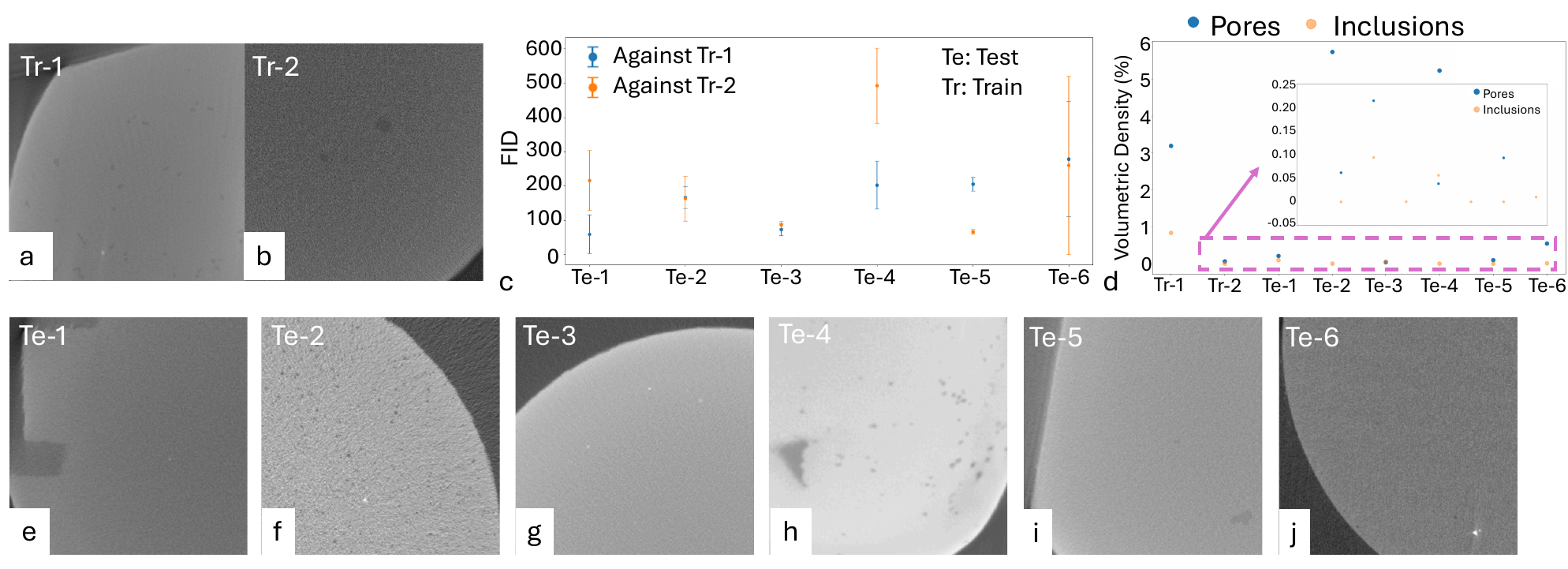}
    \caption{A summary of training (Tr-*) and testing (Te-*) data and their comparisons. GAN-generated data based on two metallic materials and different scan settings are shown in a) Tr-1 and b) Tr-2. Panels e-j show 6 crops from test data Te-1 through Te-6: Te-1: OoD test sample (Real, with more inclusions and fewer pores, and a slightly different noise distribution); Te-2: OoD test sample (Real, with more pores and fewer inclusions, and a different noise distribution); Te-3: InD test sample (with more inclusions and fewer pores, but similar noise distribution); Te-4: OoD test sample (Real, with more pores, no inclusions, and significantly less noise); Te-5: OoD test sample (Synthetic, with more pores, no inclusions, and a slightly different noise distribution); Te-6: OoD test sample (Real, with more pores, inclusions, and increased streaking noise). Panel (c) is an error-bar plot of Fréchet Inception Distance (FID) comparing test data against training data. A larger FID score indicates stronger OoD characteristics. Overall, the OoD data cover scenarios ranging from entirely different from both training datasets to being similar to one but not the other. In panel (d), we plotted pore and inclusion volume density (i.e., the proportion of the sample’s volume consisting of defects). Pore density reached up to 3\% in Tr-1, while inclusion density was under 1\%. All test data exhibit lower inclusion density between 0-0.25\%, but pore density can reach up to 6\%, making detection harder and impacting performance.}
    \label{fig:fid_ood}
\vspace{-.5cm}
\end{figure}

CycleGAN was used to generate several datasets based on different materials and X-ray CT scan settings. Two of these datasets were used for training (referred to as Tr-1 and Tr-2), and one was used as an OoD test dataset (referred to as Te-5). In addition, six real experimental test datasets were prepared. These datasets are from various metallic alloys (Al, Ti, Steel, and Ni alloys), each with different properties, and were scanned using different acquisition settings.
For each dataset, we acquired a high-quality reference scan to generate ground truth segmentation for performance evaluation. By sub-sampling the raw high-quality reference data significantly (by as much as 12×), we generated reconstructions of lower quality that included artifacts and noise. The real test datasets are referred to as Te-1, Te-2, Te-3, Te-4, Te-6, and are described in Figure~\ref{fig:fid_ood}.
While most reconstructions were performed using the standard Feldkamp-Davis-Kress (FDK) algorithm \cite{feldkamp1984practical}, for Te-4 we utilized a deep learning-based reconstruction method \cite{ziabari2023enabling}, which significantly reduced noise and provided a clean reconstruction to serve as a less-noisy OoD test dataset. It should be noted that FDK is the most widely used analytical reconstruction method, but it is known to introduce artifacts, particularly when working with sparse datasets.

We used the Fréchet Inception Distance (FID) score to compare the distribution of noise and artifacts in each type of OoD data relative to the training datasets (see Figure~\ref{fig:fid_ood}c). FID was calculated on 7 crops of size $150 \times 150$ pixels from 7 separate images for each test and training dataset. This approach avoided the impact of background and focused on differences in noise and texture distribution. Based on FID scores, these test datasets covered various OoD scenarios. The InD dataset Te-3 had the lowest FID compared to both Tr-1 and Tr-2, whereas other OoD tests exhibited different FID relationships with respect to the training data. For instance, Te-4 and Te-6 had significantly different FID values relative to Tr-1 and Tr-2, with one being quite clean and the other notably noisy. On the other hand, Te-1 and Te-5 showed inverse FID relationships with either of the training sets.

Another factor that differentiated the training data from the test datasets was the pore and inclusion volume density, i.e., the proportion of the sample's volume that consists of defects (pores and inclusions), shown in Figure~\ref{fig:fid_ood}d. Pore density reached as high as 3\% in Tr-1, while inclusion density was less than 1\%. The test datasets had lower inclusion densities, ranging from 0 to 0.25\%, although pore density could reach up to 6\%. Lower densities made detection more challenging, thus impacting model performance.
While FID scores for some test datasets may be closer to those of InD data, we ensured that the actual material properties, acquisition settings, and defect densities in the scanned objects varied significantly with respect to the training data. These differences could be further explored using domain-specific metrics, which are outside the scope of this paper.

\iffalse

% add a table for dataset with characteristics, size
%1, 2: synthetic in-dist
%3: 
\begin{table}[!htbp]
    \centering
    \begin{tabular}{|c|l|c|c|ccc|c|}
    \hline
    No. & Name   &  Type & Size & \multicolumn{4}{c|}{Characteristics} \\
     &  &  & & Artifact & Noise & Material & Reconstruction\\\hline
    D1 & AFA & GAN & - & \ding{51} & \ding{51} & Steel & FDK\\
    \hline
    
   D2 & Incl-718 & GAN & - & \ding{51} & \ding{51} & - & more defects, few inlcusions\\\hline %synthetic in-dist
    
   D3 & InD & Experimental & 4020 &  \ding{51} & \ding{55} & - &  \\\hline % test 3

   D4 & OoD, more pore, few Incl. & Experimental & 920 & \ding{51} & \ding{55} & Titanium & MBIR \\\hline   %test2
    
   D5 & OoD, no-Incl & GAN & 600 & \ding{51}& \ding{51} & - & More pores, No inclusions \\\hline % test5
    
   D6 & OoD, less pore, more Incl. & Experimental & 920 & \ding{51} & \ding{51} & - & Less pore, many inclusion \\\hline %test1
    
   D7 &  OoD, no-Incl. & Experimental & 1080 & \ding{51} & \ding{55} & - & no inclusions, only pores \\\hline %test4
    
    \end{tabular}
    \caption{Dataset specification.}
    \label{tab:dataset_desc}
\end{table}
\fi

\vspace{-0.35cm}

\section{Methodology}\label{sec:method}%\vspace{-0.35cm}
For efficient training, we leverage a parameter-efficient \emph{SAM}  \emph{Conv-LoRa} , due to its adaptability for multiclass prediction and utilization of a Mixture-of-Expert (MoE) based low-rank adaptation for fine-tuning~\cite{zhong2024convolution}. 

\noindent\textbf{Parameter Efficient Segment Anything Model (SAM) (PEFT-SAM)}

\emph{SAM}  consists of three core components: (i) a prompt encoder to enable segmentation for a given box, point, or shape in an image, (ii) an image encoder trained on various masked image datasets, and (iii) a mask decoder that captures segmentation from the embeddings obtained from the image encoder and prompt embeddings. 

\begin{figure}[!htbp]
    \centering
    \includegraphics[width=0.7\textwidth]{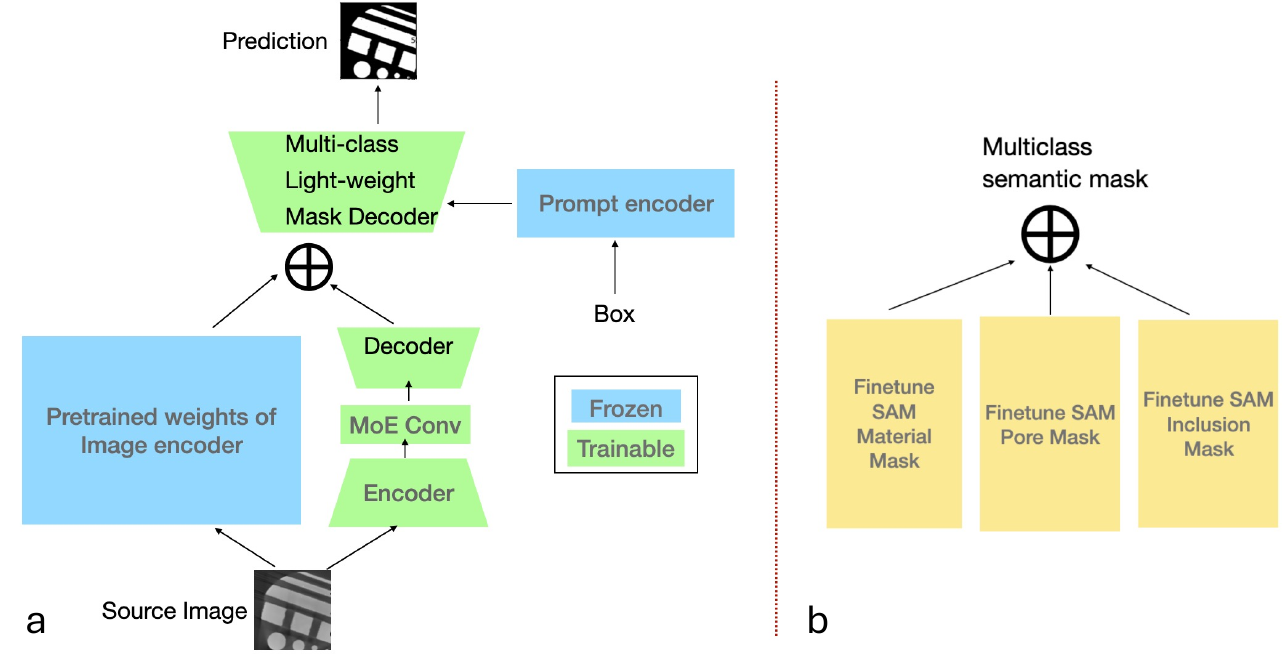}
    %\vspace{-0.25cm}
    \caption{Parameter-efficient fine-tuned \emph{SAM}  (PEFT-SAM) for multiclass segmentation. (a) Architecture of \emph{Conv-LoRa}  utilizing $3$ components of \emph{SAM}  (image encoder, prompt encoder, mask decoder). Blue denotes frozen weights of \emph{SAM} , green denotes trainable parameters for fine-tuning. The parallel encoder-decoder is the MoE-based low-rank structure that \emph{Conv-LoRa}  uses. (b) Pipeline of inference for multiclass predicted masks, where a PEFT-SAM is fine-tuned, each with a binary mask per class in XCT. Finally, the post-processing aggregator is used for multiclass segmentation.}
    \label{fig:model}
%\vspace{-0.5cm}
\end{figure}

The image encoder contains a large weight module. Mixture-of-Expert (MoE) is specifically designed to adapt a large model's fine-tuning capacity, introducing minimal computational overhead~\cite{shazeer2017outrageously}. Instead of fine-tuning the image encoder, \emph{Conv-LoRa}  adopts a parallel encoder-decoder structure alongside the frozen pre-trained weights of the image encoder. Inside this structure, \emph{Conv-LoRa}  uses a lightweight convolution operations on the encoder-decoder module as a gating mechanism, managed by MoE (expert size $n=8$). For each expert $E_i$, the embedding $H$ of the parallel structure consists of:
\[
H = \mathbf{W}_0 x + \mathbf{W}_D \sum_i^n Conv((\mathbf{W}_E x)_i N_i W_E(x)),
\]
where $N_i$ is the $i^{th}$ MoE expert module and $Conv$ is the gating mechanism for expert selection. $W_E, W_D$ are the trainable parameters of the encoder-decoder. The aggregated embedding $H$, combined with the embedding from the pre-trained image encoder, is then used to train the mask decoder. The purpose behind leveraging \emph{Conv-LoRa}  is determining scale of feature maps as for introducing different local priors to improve on performance over OoD data. As MoE consists of multiple $N$ expert networks a gating module to automatically select which expert to activate in forward propagation.
As our goal is multi-class segmentation, the target is the entire input image, where each pixel's class must be identified. To automate \emph{SAM}  without requiring explicit prompts, the prompt encoder is also frozen during training. Full fine-tuning is applied to the mask decoder, as it is a lightweight module.
For fine-tuning, the model uses a structure loss (a combination of frequency-weighted IoU loss and binary cross-entropy loss) for 15 epochs until convergence. The validation metric is IoU.

\noindent\textbf{Training:} We utilized synthetic GAN-generated data for fine-tuning. As each class (material, pore, and inclusion) is heavily imbalanced, we found that fine-tuning PEFT-SAM for each class separately led to faster convergence and better performance for OoD. To handle the class imbalance, we trained three PEFT-SAM models for three types of binary semantic masks (one for each class). Similar to the original work of PEFT-SAM \emph{Conv-LoRa} ~\cite{zhong2024convolution}, for this experiment we choose number of experts $8$ and number of trainable parameters $4.02M$ ($0.63\%$ of the pretrained SAM parameters). The number of trainable A post-processing step was then used to aggregate the predicted binary masks, resulting in a final multiclass predicted mask. Figure~\ref{fig:model}(a) shows the architecture of \emph{Conv-LoRa} , and Figure~\ref{fig:model}(b) shows the pipeline for multiclass prediction.

\vspace{-0.25cm}

\section{Experiments}\label{sec:exp}%\vspace{-0.25cm}

We aim to address the following research questions regarding the fine-tuning performance of foundational models for material microstructure segmentation:

\noindent(\textbf{R1}) How well does fine-tuning \emph{SAM}  compare to the state-of-the-art supervised models?

\noindent(\textbf{R2}) Does including generative model synthesized data improve performance?

\noindent(\textbf{R3}) Can the fine-tuned \emph{SAM}  generalize for out-of-distribution (OoD) data?

\noindent(\textbf{R4}) Does catastrophic forgetting due to fine-tuning impact performance on OoD data?

%\vspace{-0.25cm}
\subsection{Baseline}%\vspace{-0.25cm}
As the baseline, we used the 2.5D U-Net Model proposed in~\cite{cheniour2024mesoscale}. 
The model accepts and processes multiple input channels to capture 3D spatial information without the high computational cost of full 3D segmentation. Specifically, the 2.5D U-Net model processes $5$ consecutive image slices as input channels to segment one image slice at a time, leveraging the context of adjacent slices in the 3D volume.
For training, we used a single pair of CycleGAN-generated data that included both pores and inclusions. This data was augmented and preprocessed into 20,000 image patches of size $5\times256\times256$, with each patch containing five channels corresponding to five consecutive slices from the 3D volume. 80\% of the data was used for training, and 20\% for validation.
These patches were fed into the 2.5D U-Net model to capture the 3D structure of the defects while maintaining the computational simplicity of a 2D segmentation framework.

\noindent\textbf{Weighted Dice Loss Function:}
One of the key challenges during training was the extreme class imbalance between the background material, pores, and inclusions.
In fact, in Tr-1, 3\% and 0.06\% of the volume contained pores and inclusions, while in Tr-2, only 0.8\% of the volume contained pores, with no inclusions. 
In the initial stages, we employed a standard Dice loss function, commonly used in segmentation tasks due to its effectiveness in handling imbalanced data. However, this approach did not yield satisfactory performance on the validation set, likely due to the severe imbalance between the large background and the much smaller regions containing pores and inclusions.

To address this, we implemented a \textit{weighted Dice loss function}, which applies different weights to each class based on their frequency in the training data. 
The Dice loss function is defined as:
\( L_{\text{Dice}} = 1 - \frac{2 \sum_i w_i \cdot p_i \cdot g_i}{\sum_i w_i \cdot (p_i + g_i)} \),
where \( p_i \) and \( g_i \) represent the predicted and ground truth binary masks for class \( i \), and \( w_i \) is the weight for class \( i \). 
Initially, the weights \( w_i \) were calculated based on the ratio of voxels belonging to each class in the training data. 
However, this approach did not result in optimal performance, so we empirically fine-tuned the weights using a small subset of the data. 
This empirical adjustment of the weights improved the model's performance on the validation set.

\begin{figure}[!htbp]
\centering
\includegraphics[width=0.98\textwidth]{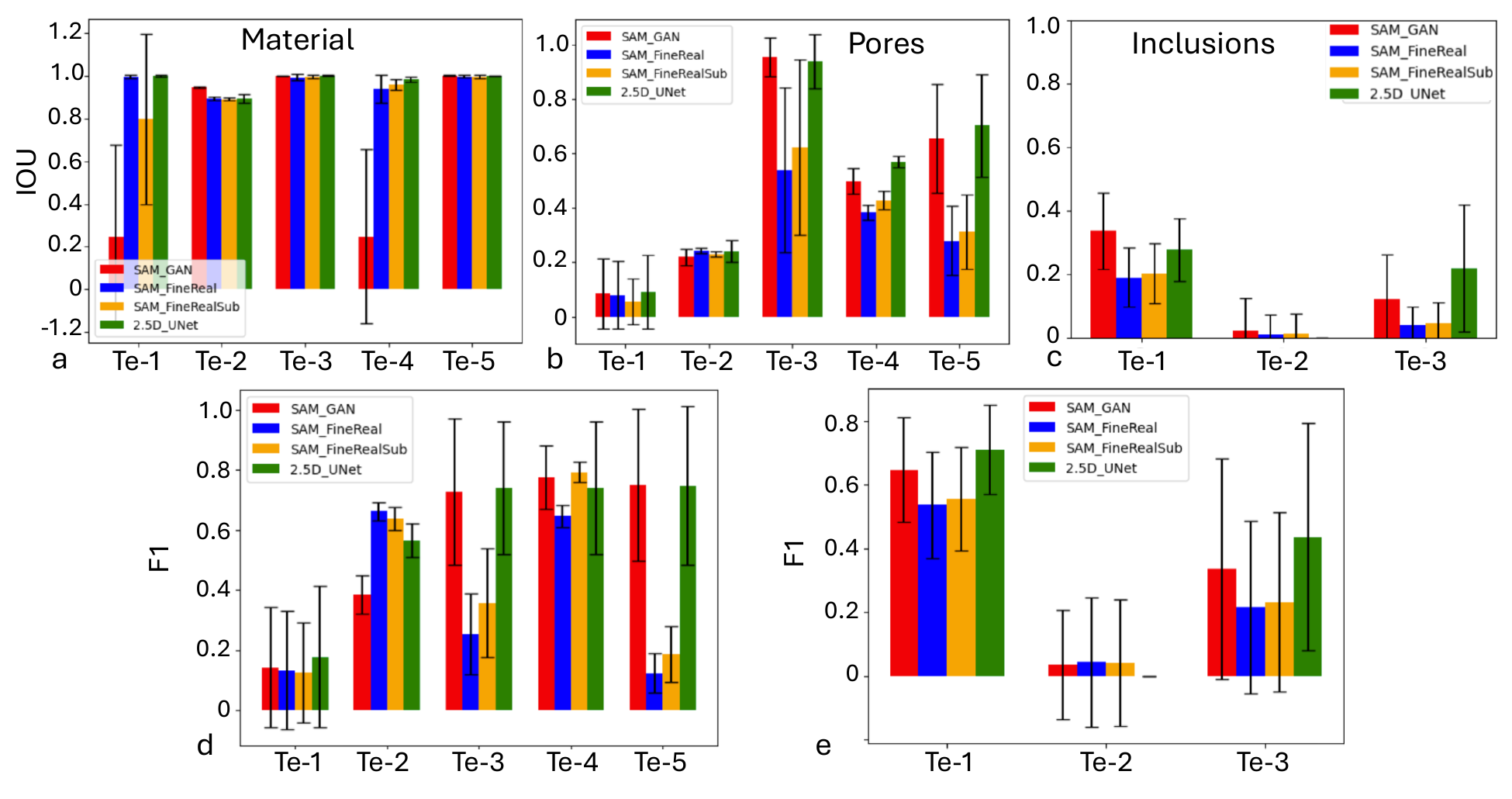}%new-evaluation/iou-all
\caption{IoU scores of \emph{SAM} fine-tuned models compared to the baseline on test datasets for (a) material, (b) pores, and (c) inclusions. Error bars (black) represent the standard deviation. The x-axis shows the test OoD data (from left to right, Te-1--Te-5). The IoU performance of \emph{SAM-GAN}  on all 3 classes is inversely proportional to the high FID score of the OoD data (shown in Fig.~\ref{fig:fid_ood}(c)). \emph{SAM-GAN}  tends to show low IoU for OoD datasets with high FID relative to the training data. Higher IoU is better. % Re-finetuning on real data (\emph{SAM-FineReal-Sub} ) tends to improve IoU by up to $1.25\%$ from the baseline.
Performance in terms of mean F1-score for (d) pores and (e) inclusions is analogous to panels (a)-(c). \emph{SAM-GAN}  outperforms the baseline on InD and OoD datasets with no inclusions. Similar to panels (a)-(c), \emph{SAM-GAN}  shows a higher F1-score for pores when the FID is low, and for inclusion, when volume of inclusions are high in the test. A higher F1-score is desired.}
\label{fig:iou_material_pore_incl}
%\vspace{-0.25cm}
\end{figure}

\noindent\textbf{Training Process:}
The 2.5D U-Net model was trained using the \textit{Adam optimizer} with an initial learning rate of \(2 \times 10^{-4}\). The learning rate was reduced whenever the validation loss stagnated for 15 consecutive epochs, with the model's parameters reset to the best-performing state to avoid overfitting. The model was trained for 150 epochs, and the best model, selected based on the highest validation Dice score, was used for inference and evaluation in the subsequent stages.

\begin{figure}
    \centering
    \includegraphics[width=.8\linewidth]{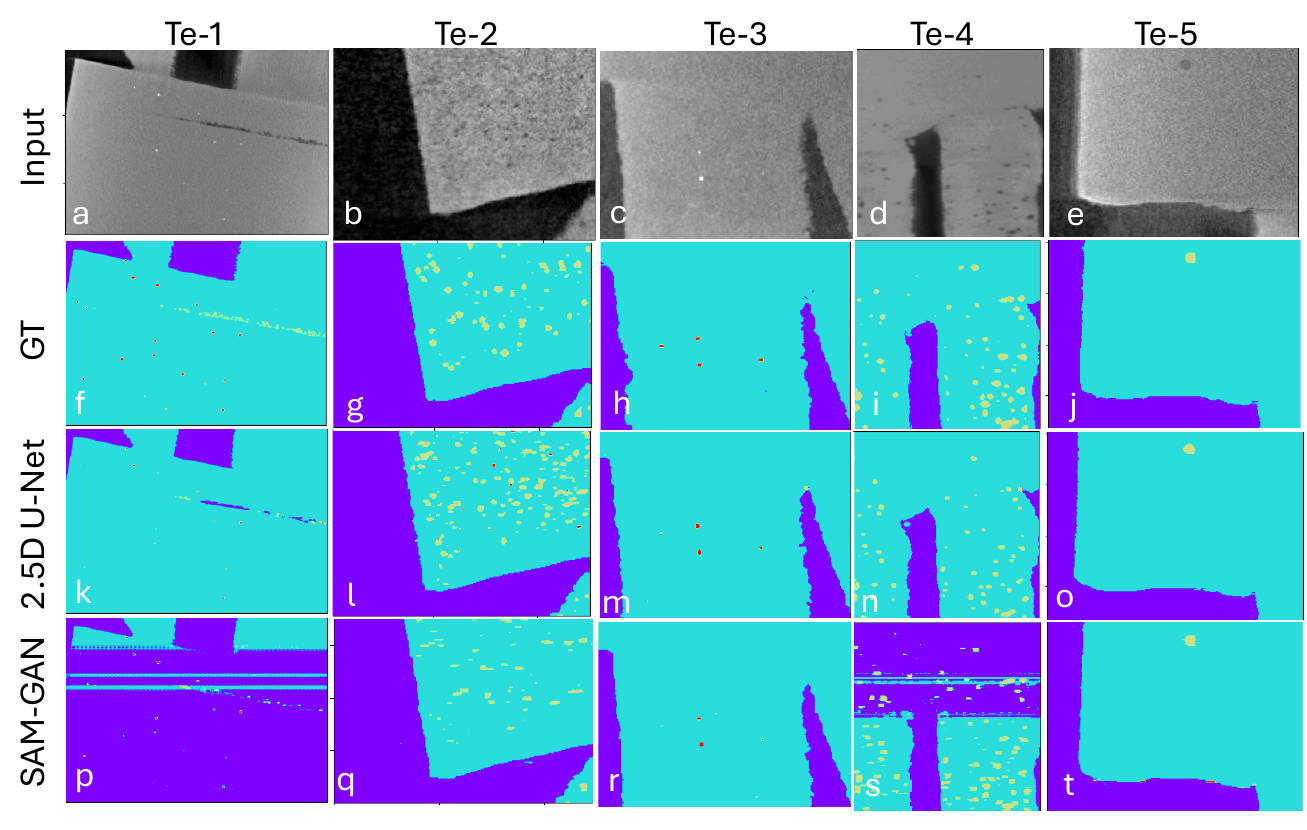}
    \caption{Performance of fine-tuned \emph{SAM} based on GAN data on InD (Te-3) and OoD tasks (Te-1,2,4,5). Each column corresponds to the test number, and each row shows the input XCT image, ground truth (GT), and the output of the fine-tuned 2.5D U-Net and SAM models (both trained on GAN-generated data). While the fine-tuned model clearly captures materials, pores, and inclusions for InD and weak OoD cases, it struggles to recognize smoother material regions in strong OoD tasks. However, the \emph{SAM} model remains consistent in detecting pores and inclusions across the different InD and OoD datasets.}
    \label{fig:finetune_gan_zero_shot}
%\vspace{-.6cm}
\end{figure}

\vspace{-0.25cm}%\vspace{-0.25cm}
\subsection{Training Data} To analyze \emph{SAM}'s robustness with Out-of-distribution (OoD) tasks, we fine-tuned \emph{SAM}  using three sets of training data. 
(i) \emph{SAM-GAN} : fine-tuned with synthetic GAN-based data from \emph{Tr-1} and \emph{Tr-2} (detailed in Sec.~\ref{sec:data}).
(ii) \emph{SAM-FineReal} : \emph{SAM-GAN}  was re-fine-tuned with $15$ experimental XCT images. For the two highest FID OoD real test datasets (Te-4 and Te-6), we selected images to add to the training data for further fine-tuning. We selected $6$ images from a sample with the same material as Te-4, scanned under the same settings and reconstructed using the same algorithm. Additionally, we selected $9$ images from the Te-6 volume and removed them from the test set for use in training. These $15$ images were used to re-fine-tune the trained SAM-GAN model.
(iii) \emph{SAM-FineReal-Sub} : \emph{SAM-GAN}  was re-fine-tuned with only the last $9$ experimental XCT images. For (ii) and (iii), we aim to analyze \emph{SAM}'s performance after further fine-tuning on a few samples of experimental data. The hypothesis is that we can improve the performance of the already trained SAM model by adding just a few real data images in multiple pairs (case ii) or a single pair (case iii).
\vspace{-0.35cm}
\subsection{Layer-by-Layer Precision, Recall, and F1-Score Calculation}%\vspace{-0.25cm}

To evaluate segmentation performance, we calculated precision, recall, and F1-score on a layer-by-layer basis for each 3D volume. 
We used this approach to align with the SAM model, as it operates on 2D slices of the 3D volume, one at a time.
For each layer \( z \), the true positives (TP) represent the correctly predicted defects that overlap with the ground truth, while false positives (FP) are defects predicted but not present in the ground truth, and false negatives (FN) are the ground truth defects missed by the predictions.
The precision for each layer is defined as the ratio of true positives to the sum of true and false positives: \( \text{Precision}_z = \frac{\text{TP}_z}{\text{TP}_z + \text{FP}_z} \). Recall is the ratio of true positives to the sum of true positives and false negatives: \( \text{Recall}_z = \frac{\text{TP}_z}{\text{TP}_z + \text{FN}_z} \). The F1-score, which balances precision and recall, is the harmonic mean of the two: \( \text{F1}_z = 2 \cdot \frac{\text{Precision}_z \cdot \text{Recall}_z}{\text{Precision}_z + \text{Recall}_z} \).

We computed these metrics for each layer and averaged them to obtain the mean precision, recall, and F1-scores across all layers: \( \text{Mean Precision} = \frac{1}{n} \sum_{z=1}^{n} \text{Precision}_z \), \( \text{Mean Recall} = \frac{1}{n} \sum_{z=1}^{n} \text{Recall}_z \), and \( \text{Mean F1} = \frac{1}{n} \sum_{z=1}^{n} \text{F1}_z \), where \( n \) is the total number of layers. We also calculated the standard deviation of these metrics across layers to assess performance variability.
To address the research questions posed in Sec.~\ref{sec:intro}, we evaluate and discuss the impact of GAN-generated data (\emph{Tr-1, Tr-2}) and improvements in segmentation and defect detection tasks on different fine-tuning variants of \emph{SAM}  and \emph{U-Net}.
For the rest of the paper, we use "experimental/real" and "synthetic/GAN" interchangeably.

\vspace{-0.25cm}
\subsection{Out-of-distribution (OoD) Performance}%\vspace{-0.25cm} 
For research questions, \textbf{(R1)} - \textbf{(R2)}, we analyzed the performance of the baseline and fine-tuned variants of \emph{SAM}  on the Out-of-distribution (OoD) datasets Te-1--Te-5 (detailed in Fig.~\ref{fig:fid_ood}). 
Fig.~\ref{fig:iou_material_pore_incl}(a)-(c) shows the IoU performance of \emph{SAM}  on different training sets and the baseline for InD and OoD test datasets (Te-1-Te-5) for the material, pore, and inclusion classes. We observed that \emph{SAM-GAN}  achieved higher IoU for InD (Te-3) for the material and pore classes, but consistently shows lower IoU for OoD compared to the baseline. This can be explained by comparing the FID score of test OoD data as mentioned in Fig.~\ref{fig:fid_ood}(c). \emph{SAM-GAN}  tends to show lower IoU for OoD data with high FID scores relative to the training data. For inclusions, comparing with Fig.~\ref{fig:fid_ood}(d), \emph{SAM-GAN}  performs better on OoD datasets with a high volume of inclusions (Te-1) and shows lower IoU for OoD datasets with a low volume of inclusions. Fig.~\ref{fig:iou_material_pore_incl}(d)-(e) shows the performance of \emph{SAM-GAN}  in terms of mean F1-score on test datasets for the pore and inclusion classes. Here, \emph{SAM-GAN}  performs comparably to the baseline for InD (Te-3) and OoD datasets with no inclusions (Te-4, Te-5), but yields lower F1-scores for OoD datasets with high FID scores (Te-1, Te-2). Similarly, \emph{SAM-GAN}  achieves better F1-scores for inclusions on OoD datasets with a high volume of inclusions and lower F1-scores for OoD datasets with low inclusion volumes. 
Fig.~\ref{fig:finetune_gan_zero_shot} shows an example of \emph{SAM-GAN}  predictions across all test datasets (Te-1-Te-5). \emph{SAM-GAN}  fails to capture the material in smoother regions of test data with high FID (Te-1 and Te-4).

\begin{figure}[!htbp]
\vspace{-0.25cm}
\centering
\includegraphics[width=0.7\textwidth]{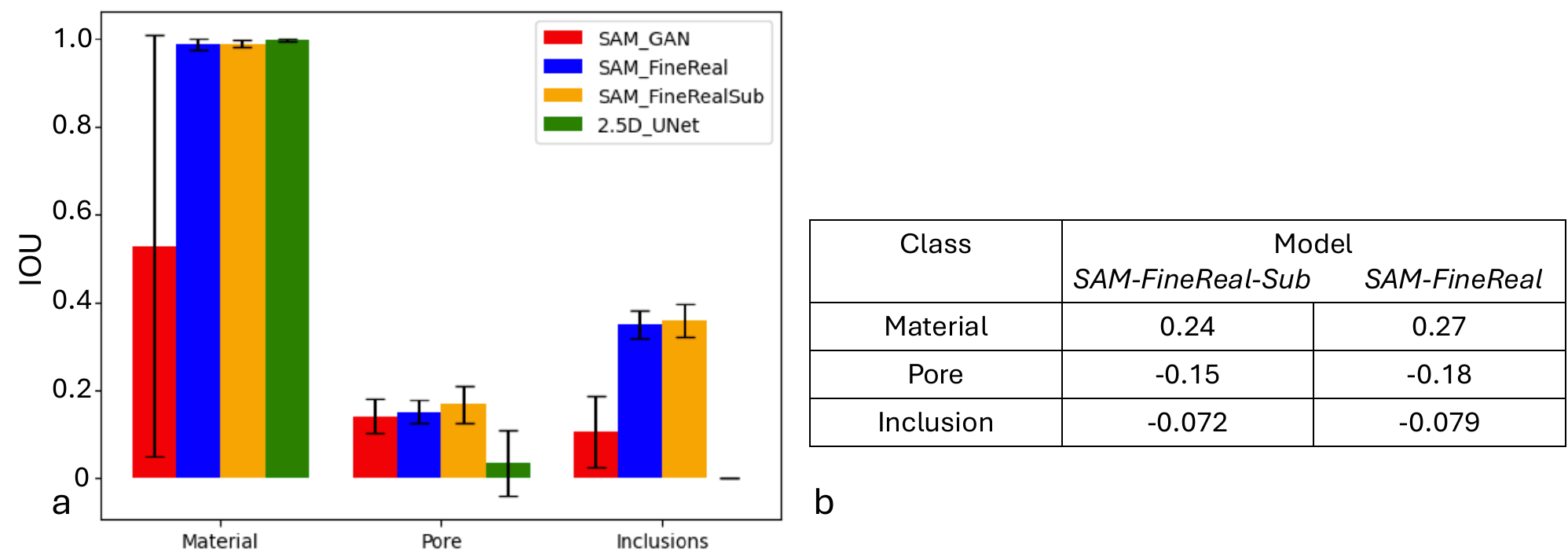}%new-evaluation/iou-all
\caption{ (a) Refinetuning performance of \emph{SAM-FineReal}  \& \emph{SAM-FineReal-Sub}  on noisy Te-6 per class. \emph{SAM-FineReal-Sub}  is robust to input noise only after fine-tuning on a very small real data.
(b) The mean change in Iou on OoD \& InD data from SAM-GAN to refinetuned models.
}
\label{fig:catastrophic_forget}
\vspace{-0.25cm}
\end{figure}

\noindent\textbf{(R3) Robustness to Noise:} 
To analyze how finetuning variants of \emph{SAM}  responds to noise, we choose to analyze the performance on a very different training noise distribution, Te-6 (high FID w.r.t both training Tr-1 and Tr-2). Figure~\ref{fig:catastrophic_forget}a shows IoU of the models for material, pore, and inclusion. 
We observe, \emph{SAM-GAN}  under-performs for all three classes. However, \emph{SAM-FineReal-Sub}  after refinetuning on a small real data increase IoU to almost $85\%$ for material, $14\%$ for pore. 

 % 'r' means right side, and width is 40% of text width
%\noindent\textbf{Ronustness to Noise:}

%\subsection{Performance on Few-shot Tasks}

%\noindent\textbf{Approach:} 

%\noindent\textbf{Results:} 
% Results after finetuning step 2
%1. Table F1 score (combined class), pore, material, inclusion ious score (combined), material, pore, inclusion

%2. Qualitative comparison plot for 5 datasets

%appendix: precision, recall values of material, pore, inclusion, combined

\noindent\textbf{(R4) Catastrophic Forgetting:}
The goal of PEFT \emph{SAM} ,\emph{Conv-LoRa}  is to leverage low rank MoE is also to mitigate catastrophic forgetting (\emph{CF}) for fine-tune training~\cite{kemker2018measuring}. We intend to further analyze this refinetuning performance of \emph{SAM-FineReal}  and \emph{SAM-FineReal-Sub}  and their change in performance on addition of new data. Figure~\ref{fig:catastrophic_forget}b shows average iou drop on all test datasets on \emph{SAM-FineReal-Sub}  and \emph{SAM-FineReal}  (2-step fine-tuning) from one step fine-tuning \emph{SAM-GAN}  for both pores and inclusion. Further, this performance drop on \emph{SAM-FineReal}  is higher than \emph{SAM-FineReal-Sub}  for $10.5\%$ on pore and inclusion. Note that, \emph{SAM-FineReal}  is fine-tuned on $9$ Te-4 data than \emph{SAM-FineReal-Sub} . 
This clearly shows presence of catastrophic forgetting with additional data and further training.

\vspace{-0.4cm}

\section{Discussion}\label{sec:discussion}%\vspace{-0.3cm}
To summarize our observations, as per research questions, for (\textbf{R1}), we observe SAM-GAN has significantly improved $11.97\%$ IoU and $33.45\%$ F1 compared to baseline (2.5D U-Net trained with same GAN-generated data) on all test OoD data across all three classes. 
For (\textbf{R2}) and (\textbf{R3}), we see training on GAN generated data can significantly improve the performance for in-distribution. However, SAM-GAN yields high performance (in terms of popular metrics) on OoD with low FID score and low performance on OoD with high FID (Fréchet inception distance) w.r.t training data. For noisy OoD (Te-6), \emph{SAM-GAN}  struggles to distinguish material class even on smoother region. 
Whereas, \emph{SAM-FineReal-Sub}  and \emph{SAM-FineReal}  with re-finetuning can improve the performance on OoD where SAM-GAN struggles. For (\textbf{R3}), although \emph{SAM-FineReal-Sub}  improves performance for OoD than \emph{SAM-GAN} , it again decrease on in-distribution, where \emph{SAM-GAN}  performance are high. Hence, results in catastrophic forgetting.

\vspace{-0.1cm}
\noindent\textbf{Challenges:}
Finetuning SAM on material XCT also pose some challenges. (i) Without a parameter efficient technique, training is computationally expensive and takes longer iteration to converge.
(ii) The lightweight mask-decoder for PEFT-\emph{SAM}  can adapt for multiclass segmentation by adding a classification layer. However, our observation shows, this cannot still adapt for manufacturing data to distinguish different classes in one image. Hence, we finetune as binary segmentation for each class separately. (iv) Refinetuning fails to predict many layers on 3D slices for OoD, although it can distinguish the materials well on 2D images. We further need to investigate the reason behind missing slices for \emph{SAM-FineReal}  and \emph{SAM-FineReal-Sub} .

\noindent\textbf{Conclusion:}
In this work, we investigated the application of the Segment Anything Model (\emph{SAM}) for the segmentation of flaws and inclusions in industrial X-ray computed tomography (XCT) data. By leveraging GAN-generated data and parameter-efficient fine-tuning techniques like \emph{Conv-LoRa}, we demonstrated that \emph{SAM}, when fine-tuned, significantly improves segmentation performance compared to the 2.5D U-Net baseline on in-distribution (InD) data. While SAM-GAN excelled in detecting inclusions and flaws in lower-noise out-of-distribution (OoD) data, it struggles with higher-noise OoD datasets. However, re-fine-tuning \emph{SAM} (\emph{SAM-FineReal-Sub}  and \emph{SAM-FineReal} ) improved performance in more challenging scenarios, although at the cost of some accuracy on InD data due to catastrophic forgetting.
Future work will focus on addressing the challenge of catastrophic forgetting during refinetuning, especially for tasks involving both InD and OoD data. We plan to explore more advanced finetuning strategies and loss functions that improve generalization to noisy data and allows for few/zero-shot learning. Moreover, future efforts will involve enhancing SAM’s ability to handle multi-class segmentation of defects, such as flaws and inclusions, in complex XCT data, considering 3D nature of the data.

\vspace{-0.35cm}

\section{Acknowledgments}
\vspace{-0.25cm}
Research sponsored by the US Department of Energy, Office of Energy Efficiency and Renewable Energy (EERE), Advanced Materials \& Manufacturing Technologies Office (AMMTO) program under contract number DE-AC05-00OR22725, and by the Artificial Intelligence (AI) Initiative as part of the Laboratory Directed Research \& Development (LDRD) at Oak Ridge National Lab (ORNL). 
The US government retains and the publisher, by accepting the article for publication, acknowledges that the US government retains a nonexclusive, paid-up, irrevocable, worldwide license to publish or reproduce the published form of this manuscript, or allow others to do so, for US government purposes. DOE will provide public access to these results of federally sponsored research in accordance with the DOE Public Access Plan (http://energy.gov/downloads/doe-public-access-plan).

\bibliographystyle{plain} 
\bibliography{main}

\newpage
\section{Appendix}
\textbf{\emph{SAM-GAN}  Finetuning Settings:} We finetune on 4 GPUs of 80GB memory for finetuning. Learning rate set to .0003 and batch size was set to 4. Each training ran for 20 epochs. Most finetuning converge by 15 iterations. Each training finishes within 4 hours. 

\begin{table}[!htbp]
    \centering
    \begin{tabular}{|c|c|}
    \hline
    No. & Size  \\\hline
    Tr-1 & 5724\\\hline
    Tr-2 & 1142\\\hline
    Te-1 & 920 \\\hline
    Te-2 & 920 \\\hline
    Te-3 & 5000 \\\hline
    Te-4 & 1080 \\\hline
    Te-5 & 600 \\\hline
    Te-6 & 785 \\\hline
    \end{tabular}
    \caption{Dataset specification.}
    \label{tab:dataset_desc}
\end{table}

\subsection{Additional Results of OoD}
\begin{figure}[!htbp]
    \centering
    \includegraphics[width=0.95\linewidth]{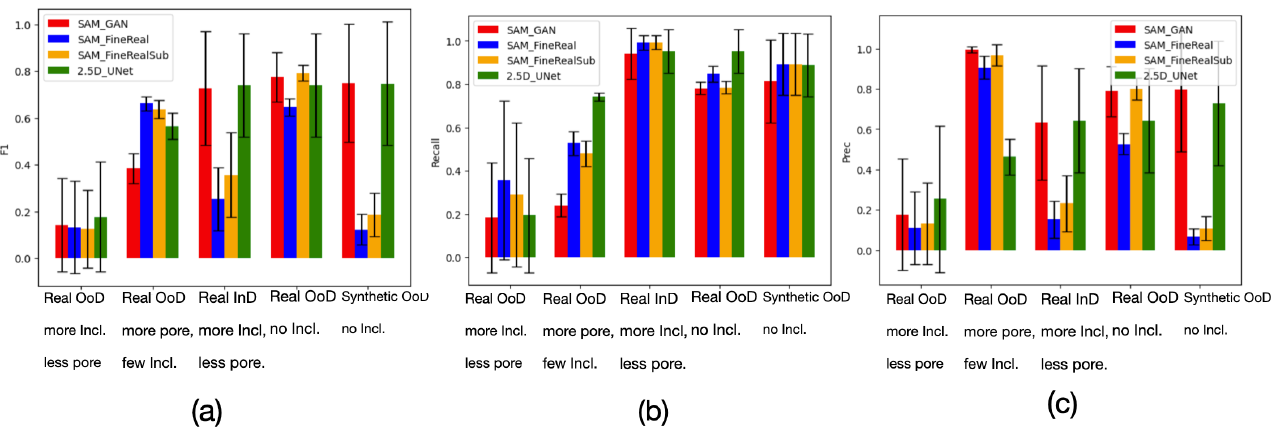}
    \caption{(a)-(c)(from left to right) mean-F1, mean-Recall, and mean-Precision values of \emph{SAM} , \emph{SAM-FineReal} , \emph{SAM-FineReal-Sub} , and \emph{U-Net}  for the class \emph{Pore}. It is noted that the recall values of \emph{SAM}  is lower for real OoD, while precisions are high in majority cases. This shows \emph{SAM}  identifies many as false negative pores than false positives in OoD.}
    \label{fig:f1_prec_recall_pore}
\end{figure}

\begin{figure}[!htbp]
    \centering
    \includegraphics[width=0.95\linewidth]{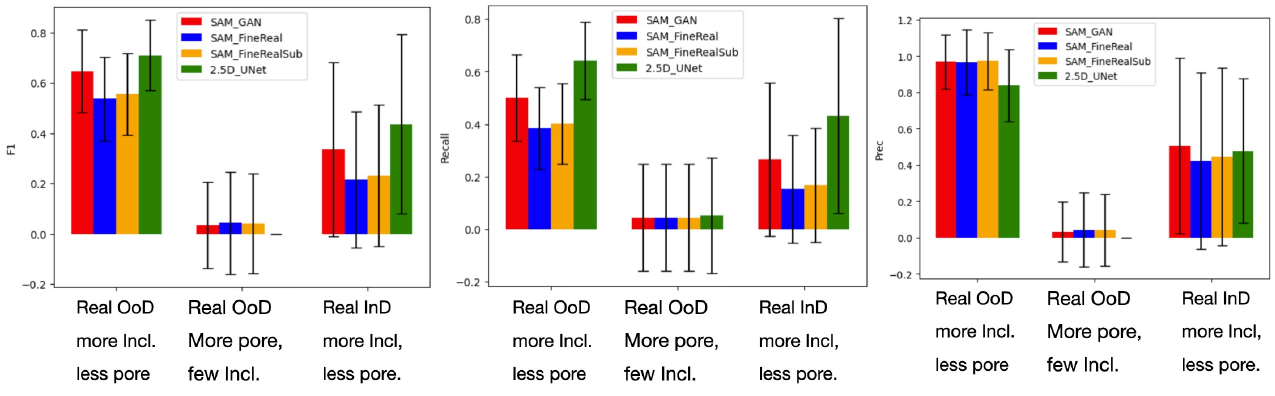}
    \caption{(a)-(c)(from left to right) mean-F1, mean-Recall, and mean-Precision values of \emph{SAM} , \emph{SAM-FineReal} , \emph{SAM-FineReal-Sub} , and \emph{U-Net}  for the class \emph{Inclusion}. It is noted that the recall values of \emph{SAM}  is lower for real OoD, while precisions are high in majority cases. This shows \emph{SAM}  identifies many as false negative pores than false positives in OoD.}
    \label{fig:f1_prec_recall_inclusiom}
\end{figure}

\begin{figure}[!htbp]
    \centering
    \includegraphics[width=0.95\linewidth]{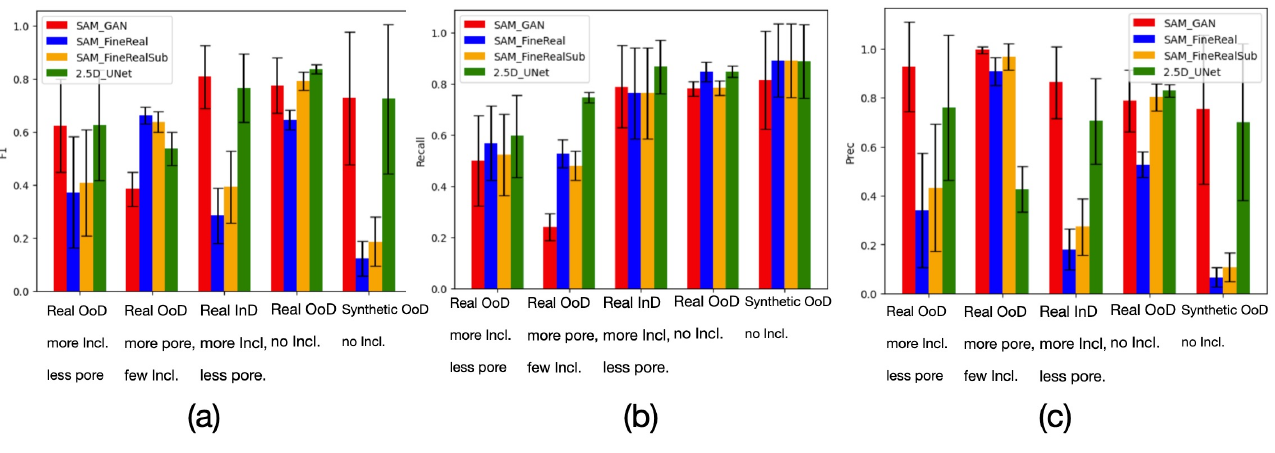}
    \caption{Analogous to Fig.~\ref{fig:f1_prec_recall_pore}, shows performance on all classes together. High value of F1 is dominated by high precision than recall.}
    \label{fig:f1_prec_recall_combined}
\end{figure}

Fig.~\ref{fig:f1_prec_recall_pore} and Fig.~\ref{fig:f1_prec_recall_inclusiom} shows the mean-F1, mean-Recall, and mean-Precision for the OoD datasets (\emph{Te-1}-\emph{Te-5}) for class \emph{Pore} and \emph{Inclusion} respectively.
FIg.~\ref{fig:f1_prec_recall_combined} shows the performance using same metric on all classes for all test OoD data.
\begin{figure}[!htbp]
\centering
    \includegraphics[width=1\linewidth]{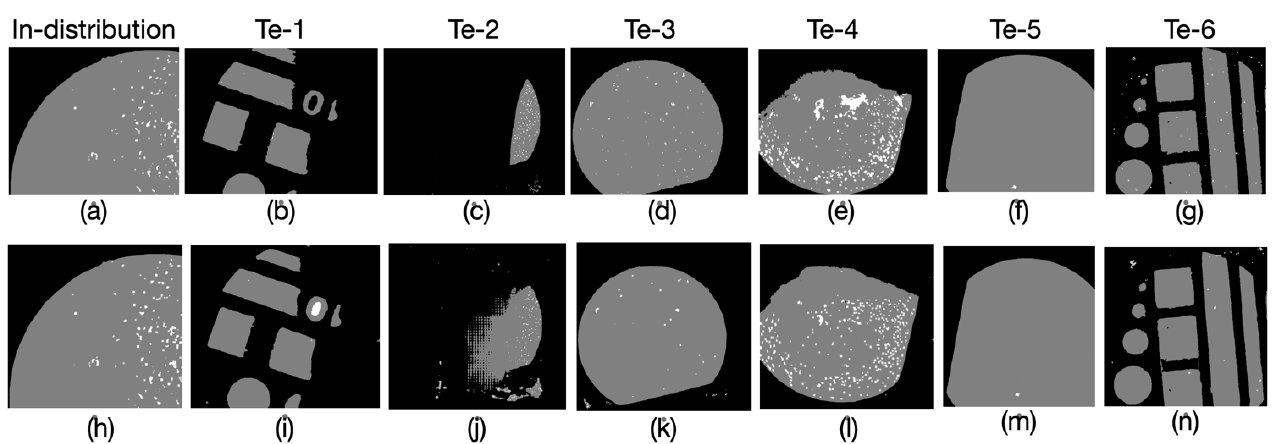}
    \caption{A sample example of the labels and corresponding predictions of the in-distribution data and all test data (Te-1--Te-6). Each column represents a dataset sample. First row (a)-(g) shows the ground-truth segmentation of material, pores, and inclusion and the second row (h)-(n) shows the corresponding prediction by Finetuned-SAM. In-distribution sample is the unseen test data similar to training Tr-1. While Finetuned-SAM can distinguish the material for all OoD, it suffers to distinguish more inclusion in Te-1 and Te-3 and dense pores Te-4 as suggested by the bar plots in Figure 4(b)-(c).}
\end{figure}

\end{document}